\documentclass[twoside,11pt]{article}

\usepackage{jmlr2e}
\usepackage{kotex}
\usepackage{color}
\usepackage{amsmath}
\usepackage{algorithm}
\usepackage{algpseudocode}
\usepackage{wrapfig}
\usepackage{svg}
\usepackage{mathtools}

\newcommand{\D}{\mathbf{D}}
\newcommand{\U}{\mathbf{U}}

\ShortHeadings{Resolution Chromatography of Diffusion Models}{Hwang, Park and Jo}
\firstpageno{1}

\begin{document}
 
\title{Resolution Chromatography of Diffusion Models}

\author{\name Juno Hwang \email wnsdh10@snu.ac.kr \\
        \addr Department of Physics Education, Seoul National University, Seoul 08826, Korea\\
       \AND
       \name Yong-Hyun Park \email enkeejunior1@snu.ac.kr \\
        \addr Department of Physics Education, Seoul National University, Seoul 08826, Korea\\
       \AND
       \name Junghyo Jo \email jojunghyo@snu.ac.kr \\
       \addr Department of Physics Education, Seoul National University, Seoul 08826, Korea\\
       Center for Theoretical Physics and Artificial Intelligence Institute, Seoul National University, Seoul 08826, Korea\\
       School of Computational Sciences, Korea Institute for Advanced Study, Seoul 02455, Korea
}


\editor{-}

\maketitle

\begin{abstract} 
Diffusion models generate high-resolution images through iterative stochastic processes. In particular, the denoising method is one of the most popular approaches that predicts the noise in samples and denoises it at each time step. It has been commonly observed that the resolution of generated samples changes over time, starting off blurry and coarse, and becoming sharper and finer. In this paper, we introduce ``resolution chromatography'' that indicates the signal generation rate of each resolution, which is very helpful concept to mathematically explain this coarse-to-fine behavior in generation process, to understand the role of noise schedule, and to design time-dependent modulation. Using resolution chromatography, we determine which resolution level becomes dominant at a specific time step, and experimentally verify our theory with text-to-image diffusion models. We also propose some direct applications utilizing the concept: upscaling pre-trained models to higher resolutions and time-dependent prompt composing. Our theory not only enables a better understanding of numerous pre-existing techniques for manipulating image generation, but also suggests the potential for designing better noise schedules.
\end{abstract}

\begin{keywords}
  diffusion models, noise schedule, resolution chromatography, upscaling, text-to-image
\end{keywords}

\section{Introduction}

In the field of image generation, diffusion-based generative models (referred to as diffusion models) have not only shown superior performance compared to other generative models such as generative adversarial networks (GANs)~\citep{dhariwal2021diffusion}, but also state-of-the-art performance in conditional sampling tasks such as text-to-image generation~\citep{Rombach_2022_CVPR}, superresolution~\citep{ho2022cascaded}, and semantic image editing~\citep{couairon2022diffedit, kwon2022diffusion}. Diffusion models learn the reverse process of slowly spreading the distribution of data to an exact prior distribution (such as a Gaussian distribution) through a neural network, and generate samples by iteratively applying the learned stochastic reverse process starting from a random initial value. 

Denoising diffusion probabilistic models (DDPMs) are currently the most widely used diffusion models. They train a neural network to predict the noise given the noised sample and the time step, in which the model gradually denoises the image during the generation process~\citep{ho2020denoising}.
Therefore, by observing the samples during the generating process at each time step, we can see that the samples gradually evolve from completely random and meaningless noises to clear and meaningful images.

After the introduction of DDPM, there is a widely observed characteristic during the generation process of a diffusion model. That is, the process begins by generating a coarse and blurry signal of the sample, which then gradually refines into finer and sharper details. 
The easiest way to measure this phenomenon is by examining the posterior $\hat x_0 = \mathbb{E}[x_0|x_t]$. Figure 1 shows the temporal evolution of $\hat x_0$ and its power spectral density (PSD), clearly demonstrating the diffusion model's coarse-to-fine behavior.
This tendency is well-known and widely utilized in various research areas such as loss design \citep{choi2022perception, hoogeboom2023simple, chen2023importance}, image editing \citep{park2023understanding}, customization \citep{daras2022multiresolution}, text-to-3D \citep{lin2023magic3d, chen2023fantasia3d, wang2023score}, and more.

Despite the widespread observation of coarse-to-fine behavior in diffusion models, we still lack a clear mathematical understanding of why and how the resolution changes during the image generation process and what factors influence it.
In this paper, we propose a mathematical analysis of the resolution change by expanding the sample across multiple resolutions and show that downsampling (coarse graining) is equivalent to the time adjustment of DDPMs. Through this, we introduce the concept of  \textit{resolution chromatography} to represent the generation rate of signals at each resolution and show that it is predetermined by the noise schedule. By employing resolution chromatography, we gain a deeper insight into the coarse-to-fine behavior of diffusion models and the role of noise schedules, allowing us to interpret previous studies' 
time-dependent techniques. Our key contributions are summarized as follows:

\begin{itemize}
\item We develop a concept of resolution chromatography that indicates the signal generation rate of each resolution, thereby elucidating the coarse-to-fine behavior of generation process in diffusion models.

\item We find that the multiple resolutions can be matched through time adjustment and intensity rescaling, and experimentally confirm that the resolution chromatography in the actual sampling process follows our theory.

\item We propose that our theory can be employed to quantitatively design the process of upscaling pre-trained models or time-dependent prompt composing.  

\end{itemize}

\begin{figure}
    \begin{center}
    \centerline{\includegraphics[width=1\columnwidth]{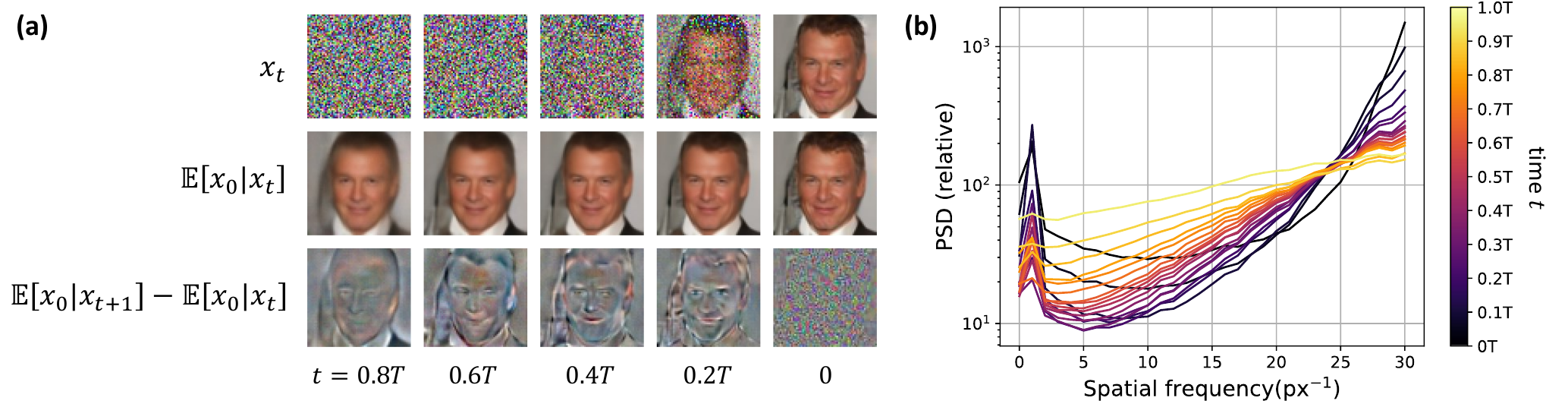}}
    \caption{Coarse-to-fine signal generation process in diffusion models. (a) Noised samples $x_t$, their corresponding denoised samples' expectation $\mathbb{E}[x_0|x_t]$, and the differences between consecutive time steps.
(b) Power Spectral Density (PSD) of changes in expectations over time, averaged across 500 samples. As time $t$ approaches to 0, the intensity in the low-frequency domain decreases, while the high-frequency domain becomes more intense, suggesting the coarse-to-fine behavior.} 
    \end{center}
    \label{fig_psd}
\end{figure}

\section{Background}
\subsection{Diffusion Models}

Diffusion models define a forward diffusion process, $x_0 \to x_1 \to \cdots \to x_T$, that starts from a noise-free original image $x_0$, and evolves to a fully diffused image $x_T$ after $T$ steps.
The final image follows a multivariate Gaussian distribution, $x_T \sim \mathcal{N}(0, I)$.

The forward progression between successive time points is delineated as a Markov process characterized by a conditional probability, $q(x_t|x_{t-1})$~\citep{sohl2015deep}.
When this Markov forward process is recursively applied, one can derive a short-cut formulation for $q(x_t|x_0)$ jumping directly from $x_0$ to $x_t$,
\begin{equation}
    x_t = \sqrt{\alpha_t}x_0 + \sqrt{1-\alpha_t} \epsilon_t,
\label{eq_diffusion}
\end{equation}
which can be interpreted as an interpolation between signal $x_0$ and noise $\epsilon_t \sim \mathcal{N}(0, I)$.

Scheduling noise corresponds to design the parameter $\alpha_t$ through time. It has been chosen heuristically. In early diffusion research, focusing on $q(x_t|x_{t-1})$ rather than $q(x_t|x_0)$, the noise schedule was defined by linearly changing. 
This schedule is called linear noise schedule. However, a following research found that most of the signal is quickly destroyed in the early stage of forward diffusion in the linear schedule, and proposed cosine scheduling to make the loss of information at each time step relatively uniform by setting $\alpha_t$ in a cosine shape~\citep{nichol2021improved}. 
It is noteworthy that the noise schedule parameter $\alpha_t$ also determines the signal-to-noise ratio (SNR):
\begin{equation}
\label{eq:snr}
    \textrm{SNR} = \frac{\alpha_t}{1 - \alpha_t}.
\end{equation}

In DDPMs~\citep{ho2020denoising}, a neural network learns the function $\epsilon(x_t, t)$ to predict the noise $\epsilon_t$ given noised signal $x_t$ and time step $t$ by minimizing the loss function,
\begin{equation}
    L = \mathop{\mathbb{E}}_{t, x_0, \epsilon_t} || \epsilon_t - \epsilon(x_t, t)||^2.
\label{eq:loss}
\end{equation} 
Then, the backward (generation) process $q(x_{t-1}|x_t)$ is modeled by $p(x_{t-1}|x_{t})$.
For clarity in notation, we differentiate noise $\epsilon_t$ and data distribution $q(x_{t-1}|x_t)$ from their corresponding models, $\epsilon(x_t, t)$ and $p(x_{t-1}|x_t)$, respectively.

To make the generation process, $x_T \to x_{T-1} \to \cdots \to x_0$, deterministic and reduce sampling steps, the denoising diffusion implicit model (DDIM) introduces a non-Markovian forward distribution that has the same marginal distribution of $q(x_t|x_0)$ with DDPM~\citep{song2020denoising}. Thus, we can still use the noise predictor $\epsilon(x_t, t)$ pre-trained in DDPM just by changing the backward process as follows\footnote{The paper actually provided a family of non-Markovian models and took zero-variance limit to make it deterministic, but here we focus on the deterministic case only.}:
\begin{equation}
x_{t-1} = \frac{1}{\sqrt{\alpha_t/\alpha_{t-1}}} x_t + \bigg( \sqrt{1-\alpha_{t-1}} - \frac{1}{\sqrt{\alpha_t/\alpha_{t-1}}} \sqrt{1-\alpha_{t}} \bigg) \epsilon(x_t, t).
\label{eq:ddim_step}
\end{equation}
This modification not only provides a better sampling quality in small sampling steps, but also allows an inversion from a given image to the initial noise $x_T$, which makes image editing coherent in many techniques~\citep{hertz2022prompt}.

\subsection{Text-to-Image Diffusion Models}
Diffusion models can work for conditional generation tasks such as generating images within specific classes.
One can achieve conditional generation by training a separate classifier and incorporating its log-likelihood gradients into the noise term $\epsilon(x_t, t)$~\citep{dhariwal2021diffusion}.
However, shortly thereafter, an alternative approach called Classifier-Free Guidance (CFG) was introduced. CFG involves training the noise predictor $\epsilon(x_t, t; c)$ with a specific condition $c$, while also incorporating a portion of training data that lacks this condition~\citep{ho2022classifier}. In the sampling process within CFG, the original noise prediction is replaced with a linear combination of both conditional and unconditional noise predictions:
\begin{equation}
\label{eq:cfg_definition}
    \tilde{\epsilon}(x_t,t;c)=
    {\underbrace{\epsilon(x_t, t)}_{\text{denoise}}}+
    w{\underbrace{[\epsilon(x_t, t; c) - \epsilon(x_t, t)]}_{\text{guidance}}}.
\end{equation}
Here $w>1$ so that it can be interpreted as an external section from unconditional to conditional noise. By adjusting $w$ experimentally, both the fidelity and faithfulness of samples to the conditions can increase. 
Given that $c$ can encompass not only basic categorical class labels but also more complex high-dimensional variables like text embeddings and images, CFG has paved the way for the advancement of text-to-image models~\citep{saharia2022photorealistic}.

Furthermore, \citet{liu2022compositional} demonstrate that we can combine multiple text prompts, i.e., $\{c_i\}_{i=1}^n$, by linearly combining each conditional noise and improve the ability to generate complex images. The composed noise prediction guided by multiple prompts is

\begin{equation}
\label{eq:eps_composed}
    \tilde{\epsilon}_{\mathrm{composed}}(x_t, t; \{c_i\}_{i=1}^n)=
    {\underbrace{\epsilon(x_t, t)}_{\text{denoise}}}+
    \sum_i^n w_i{\underbrace{[\epsilon(x_t, t; c_i) - \epsilon(x_t, t)]}_{\text{guidance}}},
\end{equation}
where $w_i$ is the weight parameter for each prompt, working like CFG strength. 

\subsection{Diffusion Models and Resolution}

The relationship between time and resolution in diffusion models has been observed in various studies.
\citet{choi2022perception} posits a hypothesis that diffusion models have crucial time steps during which significant features in images are generated. They propose a loss that accelerates learning by assigning greater weight to these particular time step ranges.
\citet{park2023understanding} observed the PSD of the latent basis, which represents the most emphasized signal by the model, finding that at small $t$, the proportion of high-frequency signals is greater, whereas at large $t$, the proportion of low-frequency signals is dominant.
\citet{hoogeboom2023simple, chen2023importance} diagnosed the problem of diffusion models struggling with high-resolution image generation as being related to the noise schedule, and suggested a noise schedule tailored to the resolution.
\citet{daras2022multiresolution}, proposing multiresolution textual inversion, observed that the larger the time step \( t \), the more textual inversion learns information at higher resolutions. Here, textual inversion is a technique for training new visual token concepts in text-to-image models.
\citet{lin2023magic3d, chen2023fantasia3d, wang2023score} introduced a curriculum that initially learns the coarse structure at large \( t \) and then refines fine detail at small \( t \), resulting in efficient text-to-3D training.

These observations suggest that time (or noise schedule) is related to the resolution in the image generation process. Specifically, they reveal that as time increases (which means SNR decreases), the contribution shifts from coarse signals to fine signals. 
More intuitive and straightforward way to observe the relation is by examining the expectation of the original signal $\mathbb{E} [x_0 |x_t]$, which represents the generated signal at the time step. This interpretation is reasonable considering the sample $x_t=\sqrt{\alpha_t}x_0+\sqrt{1-\alpha_t}\epsilon_t$ as a linear combination of the predicted original signal $x_0$ and pure noise $\epsilon_t$, giving the relation
\begin{equation}
\label{eq:x0_expectation}
    \mathbb{E}[x_0|x_t]=\frac{1}{\sqrt \alpha_t} \left[ x_t - \sqrt{1-\alpha_t} \epsilon(x_t,t) \right].
\end{equation}
Here, it becomes more visually clear when we focus on the change of generated signal over time, namely, the difference of the expectations $\mathbb{E} [x_0 |x_{t+1}]-\mathbb{E} [x_0 |x_t]$. As shown in Fig \ref{fig_psd}, as $t$ increases, the strength of the low-frequency signal of the change intensifies.

\section{Theory} 

In this section, we demonstrate how resolution and time are related mathematically.
First, in Section 3.1, we explore the relationship between different resolutions through a single downsampling and corresponding time adjustment, based on the SNR matching.
In Section 3.2, we generalize the discussion to multiple resolutions, decomposing the noise with iterative downsampling.
In Section 3.3, we propose the concept of \textit{resolution chromatography}, which indicate the relative signal generation rate of each resolution at each time step during the image generation process of a diffusion model.

\subsection{Basic Idea : SNR Matching}

DDPMs exhibit strong performance in image generation; however, generating high-resolution images remains a considerable challenge. 
Efforts have been made to address this issue by generating low-resolution images first, then using them as a basis to create higher-resolution images.
Cascaded diffusion models (CDMs) are one such approach. They generate a low-resolution image, and then upscale it to a high-resolution image using the low-resolution image as a condition ~\citep{ho2022cascaded}.

However, a simpler naive approach may be to use the given low-resolution DDPM, which shows already good performance, as a template and add a new DDPM that learns just residual high-resolution parts that the low-resolution DDPM cannot generate\footnote{The cascaded diffusion models go through conditional generation models and start from $x_T$ again when upsampling, thus repeating the entire diffusion process multiple times. However, the naive method repeats the diffusion process only once and takes the sum of multi-resolution noise predictors during the process. The detailed generation process of the naive method will be explained in Section~\ref{section:Upscaling}}. 

\begin{figure}[t]
\begin{center}
\centerline{\includegraphics[width=0.7\columnwidth]{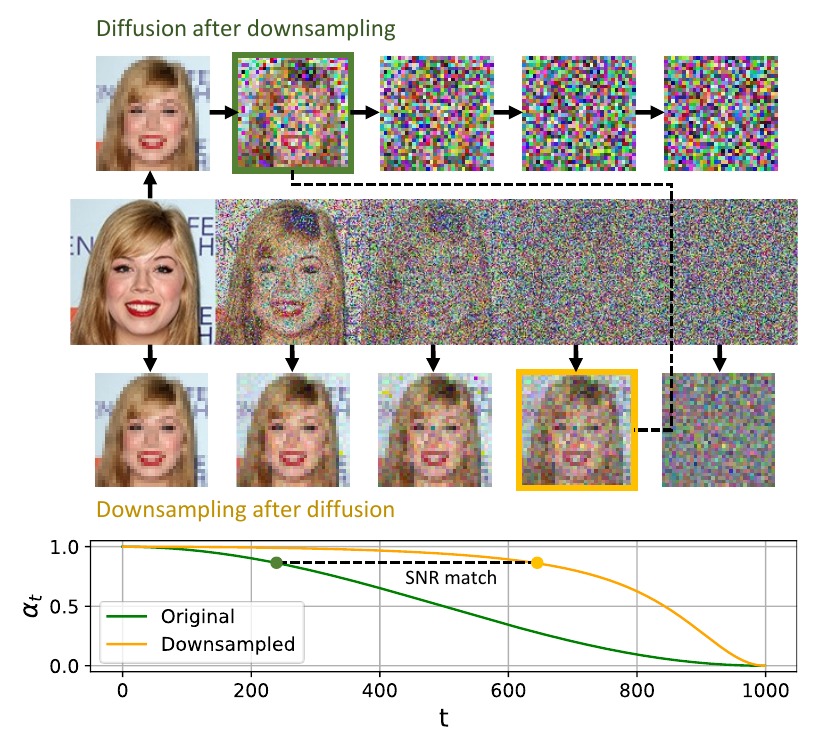}}
\caption{Time adjustment for SNR match.  
In the middle, we observe the diffusion process of a pristine, high-resolution image. At the bottom, we can see downsampled versions of these high-dimensional images after undergoing the diffusion process. The noise schedule, which dictates the signal-to-noise ratio (SNR), undergoes distinct alterations when applied to high-resolution and low-resolution images. Initially, the green-bounded image matches its SNR to that of the yellow-bounded image, which represents a downsampled compartment of the original high-resolution image, at an earlier stage of the diffusion process. In this figure, we used a kernel size of $n=4$ to emphasize  the difference in noise variance.} 
\label{fig1}
\end{center}
\end{figure}

Inspired by the cascaded generation process, we conducted a comparison between low-resolution images: (i) generated through a low-resolution diffusion process and (ii) obtained by downsampling high-resolution images generated via a high-resolution diffusion process.
Let us explain this process in detail.
First, we consider a low-resolution image, $x_0^{\mathrm{low}} = \mathbf{D}[x_0]$, where $\mathbf{D}$ represents a coarse-graining (downsampling) operator that is equivalent to the average pooling with a kernel size $n$.
In the later discussion, we use $n=2$ (e.g., 128×128 image being downsampled to 64×64) without loss of generality.
Now, we imagine the diffusion process of the low-resolution image:
\begin{equation}
    x_t^{\mathrm{low}} = \sqrt{\alpha_t} x_0^{\mathrm{low}} + \sqrt{1-\alpha_t} \epsilon_t^{\mathrm{low}}.
\end{equation}
Initially, we anticipated that the downsampled $x_t$ would align with $x_t^{\mathrm{low}}$:
\begin{align}
   \mathbf{D}[x_t] &= \sqrt{\alpha_t} \mathbf{D}[x_0] + \sqrt{1-\alpha_t} \mathbf{D}[\epsilon_t] \nonumber \\ 
   &= \sqrt{\alpha_t} x_0^{\mathrm{low}} + \frac{\sqrt{1-\alpha_t}}{2} \epsilon_t^{\mathrm{low}}.
\end{align}
The average pooling $\mathbf{D}[x_0]$ of $x_0$ does not affect the intensity.
However, we note that average pooling $\mathbf{D}[\epsilon_t]$ of $\epsilon_t$ reduces the intensity. 
When $n\times n$ pixels of Gaussian noise $\epsilon_t$ are averaged, its standard deviation is reduced by $n$ times following the central limit theorem.
We observe that both overall intensity and SNR of downsampled images $\mathbf{D}[x_t]$ after diffusion are different from diffused images $x_t^{\mathrm{low}}$ after downsampling (Figure~\ref{fig1}). 
To make them consistent, it is necessary to adjust intensity and time as 
\begin{align}
    \lambda_t \mathbf{D}[x_t] &= \lambda_t \sqrt{\alpha_t} x_0^{\mathrm{low}} + \frac{\lambda_t \sqrt{1-\alpha_t}}{2} \epsilon_t^{\mathrm{low}} \nonumber \\
    &= \sqrt{\alpha_\tau} x_0^{\mathrm{low}} + \sqrt{1-\alpha_\tau} \epsilon_\tau^{\mathrm{low}}=x_\tau^{\mathrm{low}},
\end{align}
where $\epsilon_\tau^{\mathrm{low}} = \epsilon_t^{\mathrm{low}} \sim \mathcal{N}(0, I)$.
Matching the SNRs yields the following relation:
\begin{equation}
\label{eq:snr_matching}
    \frac{4\alpha_t}{1-\alpha_t} = \frac{\alpha_\tau}{1-\alpha_\tau},
\end{equation}
and preserving the intensities gives us:
\begin{equation}
    \lambda_t^2\left(\alpha_t + \frac{1-\alpha_t}{4}\right) = 1.
\end{equation}
The scale factor $\lambda_t$ can match their overall intensities, and the adjusted time $\tau$ can match their SNR:
\begin{align}
\label{eq:scaling_relation}
    \lambda_t = \frac{2}{\sqrt{1 + 3 \alpha_t}}, \phantom{M}
    \tau = \mathrm{SNR}^{-1}\left( 4 \mathrm{SNR}(t) \right).
\end{align}
Here we treat $\mathrm{SNR}(t)$ as a function of time for simplicity in notation, and its inverse  can be derived from Equation~\eqref{eq:snr}. To have the same SNR, $\tau$ is smaller than $t$.
This observation constitutes the core discovery of our research: low-resolution signals embedded within high-resolution images exhibit a slower rate of decay in diffusion models.

\subsection{Generalization}
\begin{figure}[t]
\begin{center}
\centerline{\includegraphics[width=0.5\columnwidth]{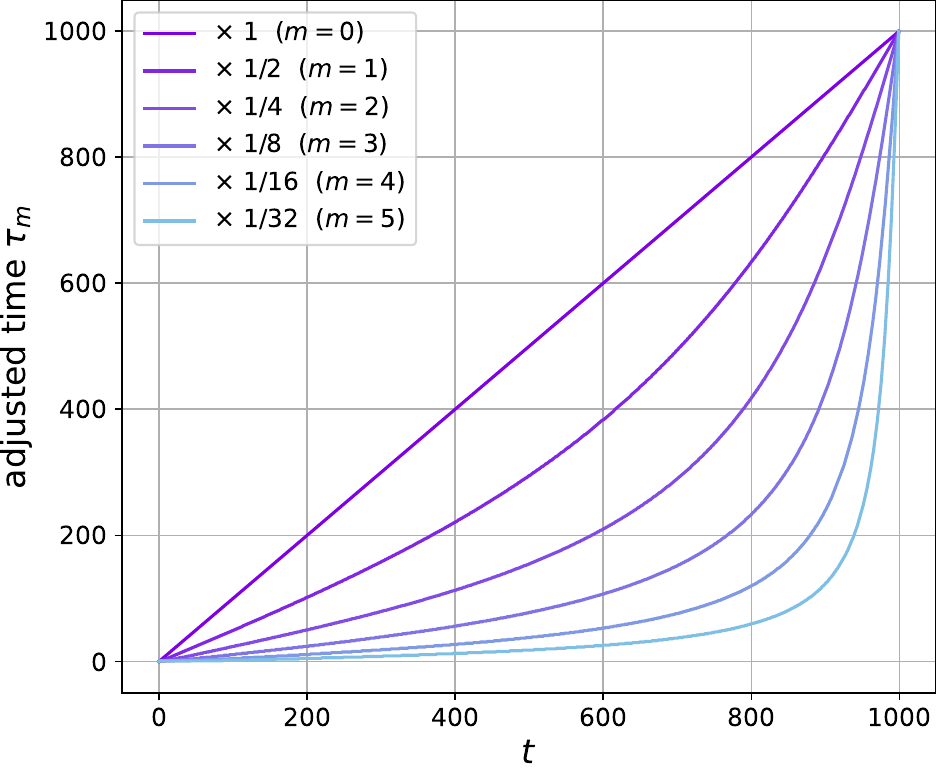}}
\caption{
Time adjustment for cosine schedule under iterative downsampling with kernel size of $n=2$ .
} 
\label{fig_tlow}
\end{center}
\end{figure}

We can extend this concept to examine lower-resolution signals. Let's denote the low-resolution image as $x_{\tau_1}^{(1)} = x_\tau^{\mathrm{low}}$, indicating a one-step downsampling with the appropriate time adjustment, $\tau_1$. Subsequently, we investigate lower-image signals $x_{\tau_m}^{(m)}$ of the $m$-th lower resolution, each adjusted with the proper time, $\tau_m$. These signals exhibit the following scaling relationship with the original resolution image $x_t$:
\begin{equation}
    x^{(m)}_{\tau_m} = \lambda_t^{(m)} \D^m [x_t].
\end{equation}
As derived earlier, the intensity factor $\lambda_t^{(m)}$ and time adjustment $\tau_m$ can be determined in a similar manner:
\begin{align}
    \lambda_t^{(m)}\mathbf{D}^m [x_t] &= \lambda_t^{(m)} \sqrt{\alpha_t} \mathbf{D}^m [x_0] + \lambda_t^{(m)}\sqrt{1-\alpha_t} \mathbf{D}^m [\epsilon_t] \nonumber \\
    &= \lambda_t^{(m)} \sqrt{\alpha_t} x_0^{(m)} + \frac{\lambda_t^{(m)} \sqrt{1-\alpha_t}}{2^m} \epsilon_t^{(m)} \nonumber \\
    &= \sqrt{\alpha_{\tau_m}} x_0^{(m)} + \sqrt{1-\alpha_{\tau_m}} \epsilon_{\tau_m}^{(m)} =  x^{(m)}_{\tau_m},
\end{align}
where $\epsilon_{\tau_m}^{(m)} = \epsilon_t^{(m)} \sim \mathcal{N}(0, I)$.
Here, the adjustment for intensity and time is generalized as
\begin{align}
\label{eq:generalized_relation}
    \lambda_t^{(m)} = \frac{2^m}{\sqrt{1 + (2^{2m}-1) \alpha_t}}, \phantom{M}
    \tau_m = \mathrm{SNR}^{-1}\left( 2^{2m} \mathrm{SNR}(t) \right).
\end{align}
The adjusted time has the following orders (Figure ~\ref{fig_tlow}):
\begin{equation}
    t=\tau_0 > \tau_1 > \tau_2 > \cdots > \tau_{m}.
\end{equation}
This temporal order implies that lower-resolution signals exhibit a slower rate of decay, while higher-resolution signals decay more rapidly. By reversing this statement for the forward process, we can infer the resolution-dependent image generation in the backward process. We make the assumption that the backward model, denoted as $p(x_{t−1}|x_t)$, accurately learns the backward process, $q(x_{t−1}|x_t)$, of the data. Notably, lower-resolution signals tend to persist to a greater extent at later time points of $t \sim T$. During the image generation from $x_T$ to $x_0$, lower-resolution signals become discernible earlier as $t$ decreases from $T$, while higher-resolution signals become apparent later as $t$ approaches $0$.

\subsection{Resolution Chromatography}
We now proceed to assess the relative contributions of different resolution signals throughout the diffusion process. During the forward process of each resolution, the noise schedule $\alpha_{\tau_m}$ serves as an indicator of the influence of signals pertaining to the specific resolution level $m$. It is important to note that we use the adjusted time parameter $\tau_m$ to investigate the contribution of each resolution at a given time point $t$.

To precisely determine when a particular signal becomes prominent, we examine the rate of change of these signals, expressed as $d\alpha_{\tau_m}/dt$. Consequently, we define the relative contributions of each resolution component as follows:
\begin{equation}
\label{eq:def_chromatography}
    r_m(t) = \frac{1}{Z} \frac{d \alpha_{\tau_m}}{dt},\phantom{M}Z=\sum_m \frac{d \alpha_{\tau_m}}{dt}.
\end{equation}
We refer to the quantity $r_m$ as {\it resolution chromatography}, which means that how much signal of the $m$-th resolution is being generated at a given time.
Given a noise schedule $\alpha_t$, the corresponding chromatography $r_m$ can be readily determined, as $\tau_m$ can be obtained using Equation \eqref{eq:generalized_relation}. 
However, when $\alpha_t$ is complicated or cannot be expressed in a closed form, this calculation becomes difficult. Here, we introduce the following theorem which helps in its practical calculation and also clearly reveals the relationship between resolution chromatography and the noise schedule.
\begin{theorem}
Let $\alpha_t$ and $\alpha_{t}'$ be two monotonically decreasing noise schedules, and $r_m(t)$ and $r_m'(t)$ their respective resolution chromatographies. Suppose there exists a mapping $t'(t)$ such that $\alpha_t = \alpha_{t'}'$. Then, for all $m$, it follows that $r_m(t) = r_m'(t')$.
\label{thm:equiv_of_chrom}
\end{theorem}
This implies that aligning the time according to the noise schedules results in a corresponding alignment of the resolution chromatographies, offering profound insights into the significance of the role of noise schedules.
This theorem proves immensely valuable as it enables the calculation of chromatography for any noise schedule simply by referencing a single standard chromatography.
Indeed, we propose a standard noise chromatography adopting the Ornstein-Uhlenbeck process, accompanied by a comprehensive explanation and proof of the aforementioned theorem in Appendix A.

\begin{figure}[!t]
    \centering
    \includegraphics[width=1\linewidth]{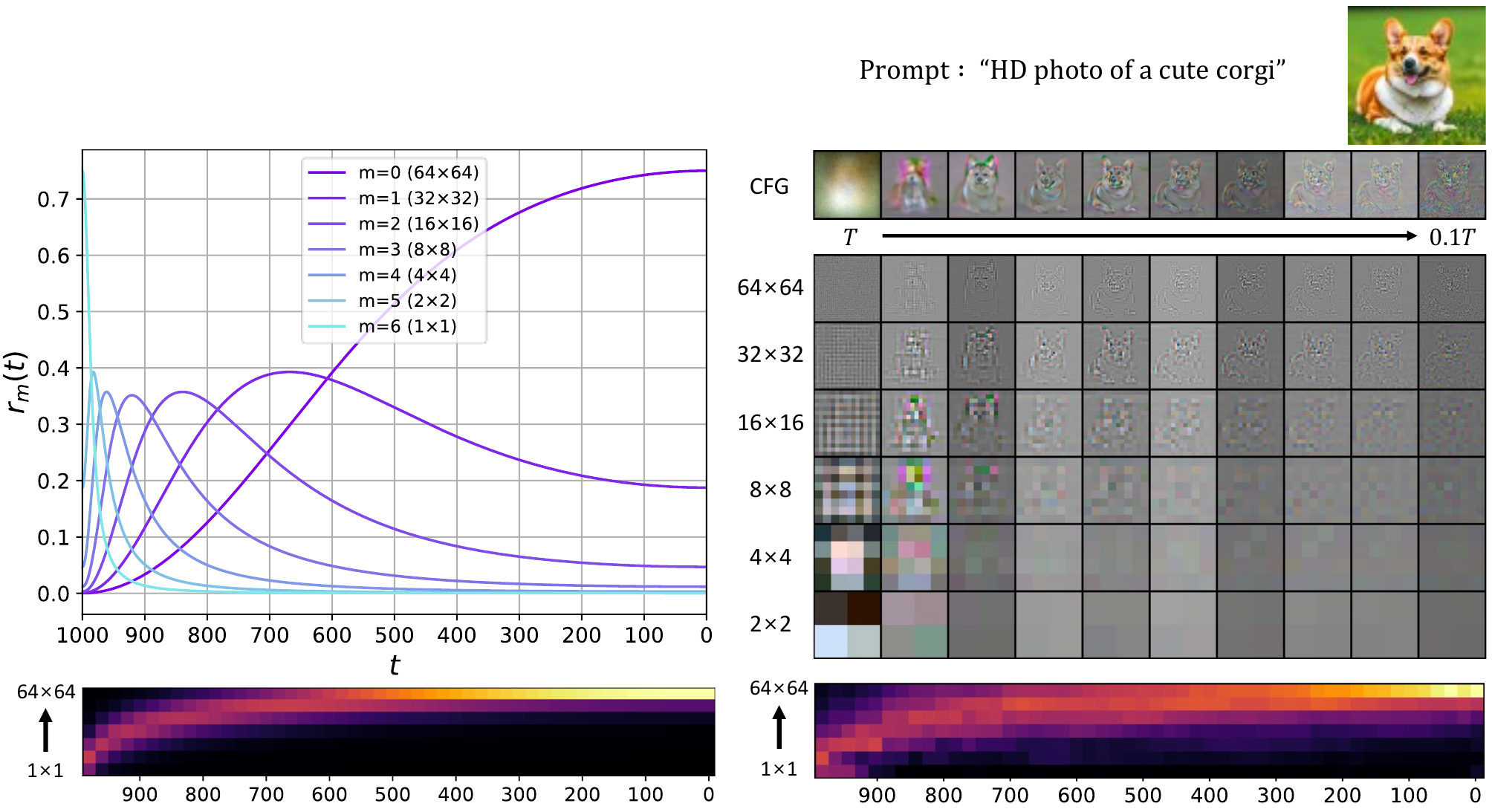}
    \caption{Resolution chromatography. 
    Left: Theoretical calculation of the evolving contribution of signals at various resolutions, denoted by $r_m(t)$, through backward diffusion processes from $t=T$ to $0$.
    Right: The decomposition of resolution chromatography for classifier-free guidance, highlighting the individual contributions of different resolutions, denoted by $\epsilon_{\textrm{CFG}}^{(m)}$. The heat maps at the bottom represent the relative intensity norm of each noise predictor.}
    \label{fig:cfg}
\end{figure}

\section{Experiment}

In this section, we verify our theory and introduce two applications inspired by our theory.
In Section 4.1, we present a method for experimentally measuring resolution chromatography in text-to-image models, along with the corresponding results.
In Section 4.2, we introduce a method for upscaling models trained at smaller resolutions using independent models trained for the residual higher-resolution noise.
In Section 4.3, we present time-dependent prompt composing, which allows for the adjustment of prompt conditions according to resolution in text-to-image diffusion models.

\subsection{Resolution Chromatography of Text-to-Image Diffusion Models}
To validate the theoretical resolution chromatography, we conducted an examination involving CFG image generation. During our investigation, we observed that the guidance term in Equation \eqref{eq:cfg_definition} can be understood as the signals related to a condition $c$, which is mathematically represented as follows:
$\epsilon_{\mathrm{CFG}}(x_t, t;c) = \epsilon(x_t, t;c) - \epsilon(x_t, t)$.

Then, in our experimental setup, which has a full resolution of $64\times64$ and downsamples with a kernel size of 2,  we extract the contribution of each resolution of the guidance $\epsilon_{\mathrm{CFG}}$:
\begin{align*}
    \epsilon_{\mathrm{CFG}}^{64\times64} &= \epsilon_{\mathrm{CFG}}^{(0)} = \epsilon_{\mathrm{CFG}} - \U\D[\epsilon_{\mathrm{CFG}}] \nonumber \\
    \epsilon_{\mathrm{CFG}}^{32\times32} &=\epsilon_{\mathrm{CFG}}^{(1)} = \U\D[\epsilon_{\mathrm{CFG}}] - \U^2\D^2[\epsilon_{\mathrm{CFG}}] \nonumber \\
    \epsilon_{\mathrm{CFG}}^{16\times16}&=\epsilon_{\mathrm{CFG}}^{(2)} = \U^2\D^2[\epsilon_{\mathrm{CFG}}] - \U^3\D^3[\epsilon_{\mathrm{CFG}}] \nonumber \\
    &\cdots
\end{align*}
In this context, we consider the CFG denoisers with varying resolutions as signals associated with their respective resolutions. Subsequently, we define the measured resolution chromatography of CFG in the following manner:
\begin{equation}
    \tilde{r}_m^{\mathrm{CFG}}(t) = \frac{1}{Z} ||\epsilon_{\mathrm{CFG}}^{(m)}||^2_2,\phantom{M}Z =\sum_m ||\epsilon_{\mathrm{CFG}}^{(m)}||^2_2.
\end{equation}
Figure~\ref{fig:cfg} illustrates these resolutions in relation to the time\footnote{In this experiment, we used the pixel-based text-to-image model, DeepFloyd/IF-I-XL-v1.0 \citep{deepfloyd_ifixl}, to easily visualize and understand the meaning of the guidance term.}. 
Remarkably, our investigations confirm the consistent alignment of the resolution chromatography of CFG with the theoretical predictions depicted in the heat map. Additional examples of measured chromatography across various text prompts are referred to Appendix C.

\subsection{Upscaling of Low-Resolution Models}
\label{section:Upscaling}

\begin{figure}[t]
\begin{center}
\centerline{\includegraphics[width=1\columnwidth]{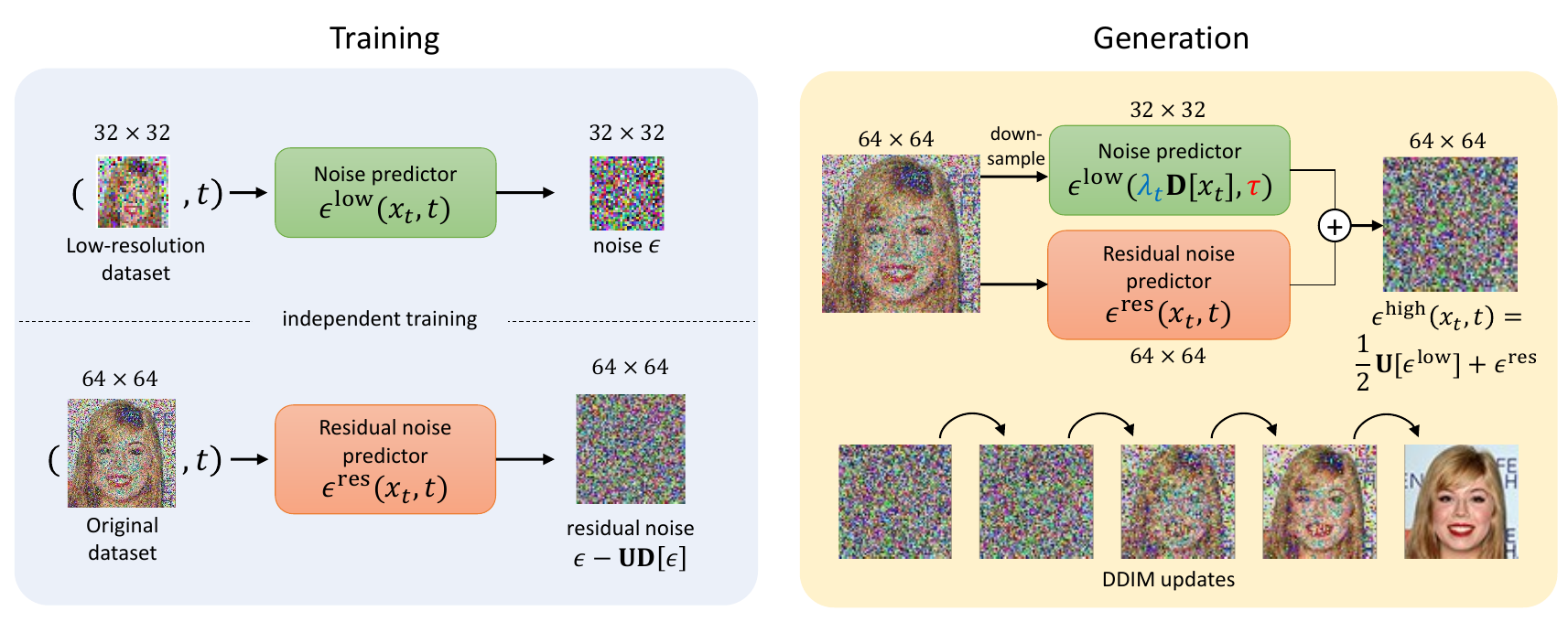}}
\caption{Cascaded image generation. Low-resolution images are utilized as templates for producing high-resolution counterparts through the integration of high-resolution residual components. The training process begins with the preparation of a low-resolution dataset, which is employed to train the low-resolution noise predictor. Subsequently, the high-resolution residual signal is independently learned through the residual noise predictor. Finally, the image generation process combines the low-resolution noise predictor and the high-resolution residual noise predictor, following appropriate intensity rescaling and time adjustments.} 
\label{fig2}
\end{center}
\end{figure}

\begin{figure}[t]
\begin{center}
\centerline{\includegraphics[width=1\columnwidth]{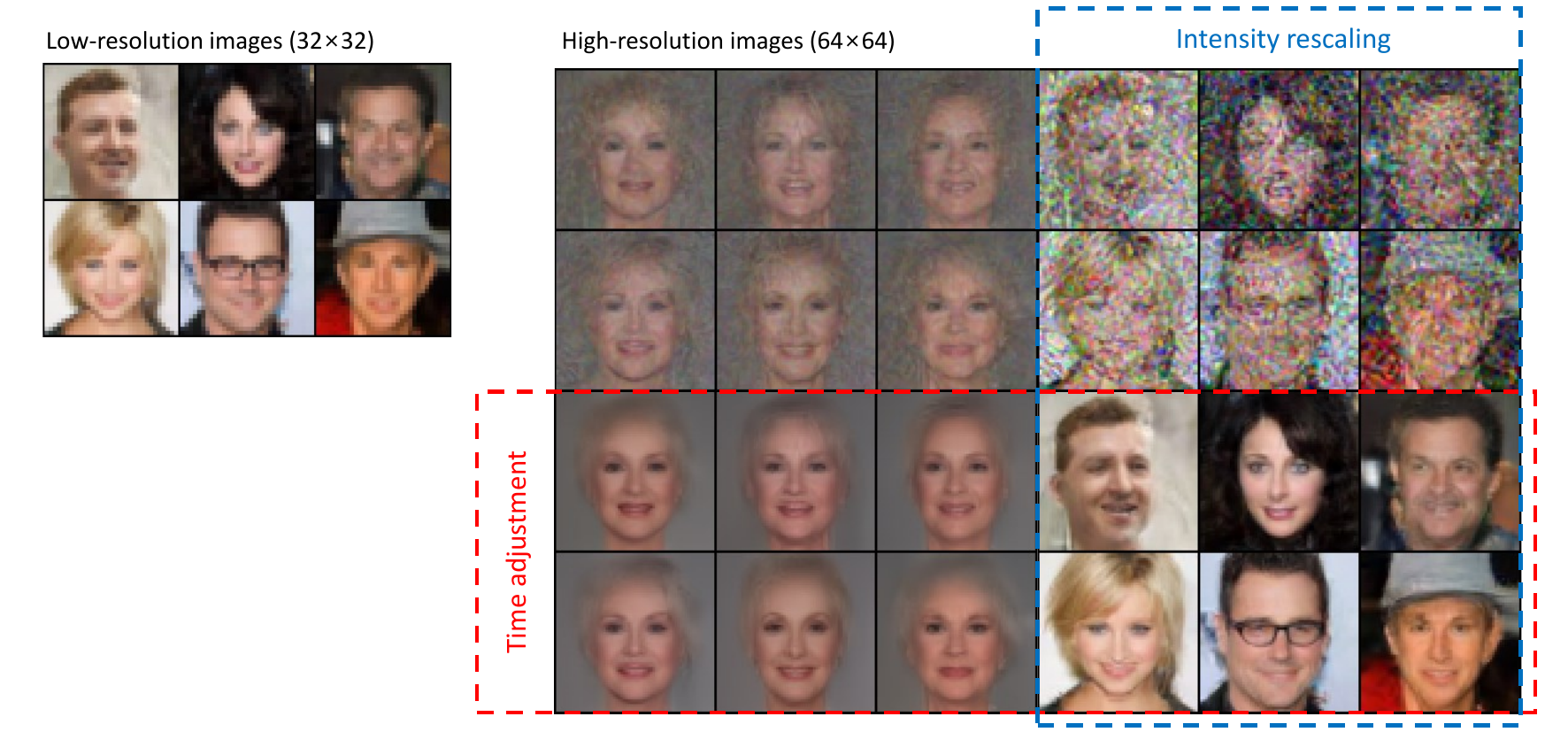}}
\caption{Time and intensity adjustment for cascaded image generation. Low-resolution images serve as templates for generating high-resolution counterparts by incorporating high-resolution residual components. The generation process incorporates intensity rescaling (indicated by blue dotted regions) and time adjustment (highlighted in red dotted regions) to ensure the successful cascaded image generation. The samples outside of dotted regions shows the results of ablation experiments for the respective adjustments.}
\label{fig3}
\end{center}
\end{figure}

Utilizing the concept of resolution chromatography, we achieve high-resolution image generation by employing a low-resolution image as the foundational template.
First, we prepare low-resolution images $x^{\mathrm{low}}_0 = \D[x_0]$ for the raw high-resolution images $x^{\mathrm{high}}_0 = x_0$.
Then, we define low- and high-resolution noises, $\epsilon^{\mathrm{low}}_t$ and $\epsilon^{\mathrm{high}}_t$, for the forward diffusion process.  
The loss for training the low-resolution noise predictor can be defined as in Equation \eqref{eq:loss}:
\begin{align}
    L^{\mathrm{low}} = \mathop{\mathbb{E}}_{t, x^{\mathrm{low}}_0, \epsilon^{\mathrm{low}}_t} || \epsilon^{\mathrm{low}}_t - \epsilon^{\mathrm{low}}(x_t, t)||^2. 
\end{align}
This loss is identical to that of the conventional DDPMs, just trained on a low-resolution dataset. In practical terms, $\epsilon^{\textrm{low}}(x_t,t)$ is a pre-trained model. 

However, when dealing with high-resolution components, rather than acquiring a dedicated high-resolution noise predictor, we train a residual noise predictor designed to capture high-resolution signals by isolating them from the signals that have emerged from low-resolution content:
\begin{align}
    L^{\mathrm{res}} &= \mathop{\mathbb{E}}_{t, x^{\mathrm{high}}_0, \epsilon^{\mathrm{high}}_t} || \epsilon^{\mathrm{high}}_t -  \U \D [\epsilon^{\mathrm{high}}_t] - \epsilon^{\mathrm{res}}(x_t, t)||^2.
\end{align}
Once we obtain the model noise predictors, $\epsilon^{\mathrm{low}}(x_t, t)$ and $\epsilon^{\mathrm{res}}(x_t, t)$, we merge them to generate high-resolution images (Figure~\ref{fig2}):
\begin{equation}
    \epsilon^{\mathrm{high}}(x_t, t) = \frac{1}{2} \mathbf{U}\big[ \epsilon^{\mathrm{low}}\big( x_\tau^{\mathrm{low}}, \tau \big) \big] + \epsilon^{\mathrm{res}}(x_t, t),
\end{equation}
where $x_\tau^{\mathrm{low}} = \lambda_t \mathbf{D}[x_t]$.
In this context, we fine-tune a low-resolution noise predictor, denoted as $\epsilon^{\mathrm{low}}( x_\tau^{\mathrm{low}}, \tau)$, to ensure it maintains the same SNR as the data at time $t$.
Additionally, we scale up the low-resolution noise predictor to match the dimension of the high-resolution noise predictor $\epsilon^{\mathrm{high}}$ by utilizing an upscaling operator $\U$ with a scaling factor of $n$, achieved through nearest-neighbor interpolation.
To ensure that the variance of $\epsilon^{\mathrm{high}}$ and $\epsilon^{\mathrm{low}}$ align, we introduce a factor of $1/2$. This adjustment becomes apparent when examining their downsampling. Specifically, $\D[\epsilon^{\mathrm{high}} - \epsilon^{\mathrm{res}}]$ exhibits a standard deviation of $1/2$, a consequence of the central limit theorem, while $\D\U\big[ \epsilon^{\mathrm{low}} \big]$ maintains a standard deviation of 1. Note that $\D\U = \mathbf{I}$, where $\mathbf{I}$ represents an identity operator, whereas $\U\D \neq \mathbf{I}$.

We achieve upscaled image generation following appropriate adjustments in both intensity and timing (Figure~\ref{fig3}), based on a $32\times32$ resolution model pre-trained on the CelebA dataset \citep{liu2015faceattributes}. 
The findings, derived from the ablation experiments concerning time adjustment $\tau$ and intensity rescaling $\lambda_t$, highlight the necessity of applying both procedures for an effective upscaling of the model.
Without time adjustment, the model fails to appropriately respond to the SNR, leading to an inability to   remove noise. Meanwhile, in the absence of intensity rescaling, although it removes noise, the signal variance is reduced, resulting in convergence to similar outcomes and a decrease in fidelity to the dataset. 
In this experiment, to improve overall quality, we employed static threshold \citep{saharia2022photorealistic} in a cascaded manner to clamp the pixel intensity of generated images, denoted as $x^{\mathrm{low}}_0$, across all resolutions within the range from -1 to 1 (refer to Appendix B for the specific algorithm and explanation for implementing  static threshold in a cascaded manner).

The scaling relation between low- and high-resolution noise predictors can be generalized.
To maximally use the idea of the scaling, let us imagine $M$ multiple virtual noise predictors, $\{ \epsilon^{(0)}, \epsilon^{(1)}, ..., \epsilon^{(M-1)} \}$, with some abuse of notation.
This is a generalization of the previous example in which the two noise predictors correspond to $\epsilon^{\mathrm{res}}=\epsilon^{(0)}$ and $\epsilon^{\mathrm{low}}=\epsilon^{(1)}$ for $M=2$.
For an example of 256×256 resolution images with $M=8$, $\epsilon^{(0)}$ is the highest resolution noise predictor, and $\epsilon^{(M-1)}$ is the lowest noise predictor with a single averaged pixel.

Let us assume that the multiple noise predictors are perfectly trained with multiple downsampled datasets, $x^{(m)}_0 = \D^m[x_0]$ by repeatedly applying $\mathbf{D}$.
Then, we train each noise predictor $\epsilon^{(m)}(x_t, t)$ to predict corresponding residual noise $\D^m[\epsilon_t] - \U^{m}\D^{{m}}[\epsilon_t]$ using the corresponding loss:
\begin{align}
    L^{(m)} &= \mathop{\mathbb{E}}_{t, x^{(m)}_0, \epsilon^{(m)}_t} || \epsilon^{(m)}_t -  \U \D [\epsilon^{(m)}_t] - \epsilon^{(m)}(x_t, t)||^2.
\end{align}
Note that for the lowest resolution $m=M-1$, the subtraction of low-resolution contribution, $\U\D[\epsilon^{(M-1)}_t]$, is absent. They can be trained in parallel, because the loss of each resolution only concerns its own dataset and noise.
Again, once we train the noise predictors for every resolution, the overall noise predictor can be decomposed as follows,
\begin{equation}
    \epsilon(x_t, t) = \sum_{m=0}^{M-1} \frac 1 {2^m} \mathbf{U}^m \big[ \epsilon^{(m)}\big(x^{(m)}_{\tau_m}, \tau_m \big) \big].
    \label{eq:epsilon_expansion}
\end{equation}
Here, $\U^m$ denotes the iteration of $\U$ operation $m$ times, while $x^{(m)}_{\tau_m} = \lambda_t^{(m)} \D^m [x_t]$ signifies the image sample at the $m$-th resolution, incorporating appropriate time adjustment $\tau_m$.
The inclusion of the coefficient $1/2^{m}$ is intended to accommodate the reduction of the noise's standard deviation by half at each downsampling step.

\subsection{Time-Dependent Prompt Composing}

\begin{figure}[!t]
    \centering
    \includegraphics[width=0.8\linewidth]{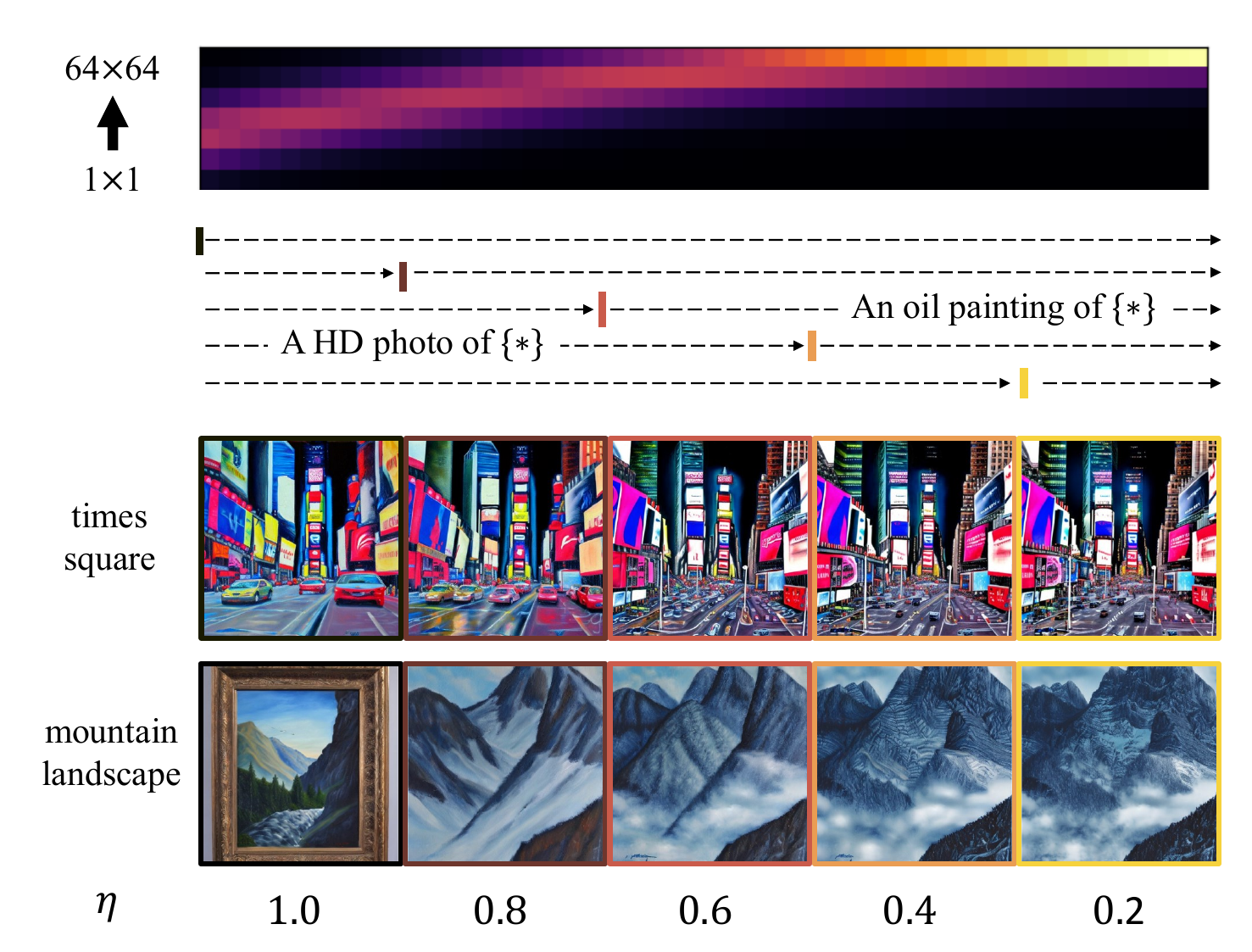}
    \caption{ Time-dependent conditional image generation and its resolution chromatography.  Two prompts, ``A HD photo of \{*\}" and ``An oil painting of \{*\}," are employed in the Stable-Diffusion-v1-5  \citep{Rombach_2022_CVPR}. The timing of their integration ($\eta T$) results in the display of final images of \{times square\} and \{mountain landscape\} at the bottom, each marked with its respective color boundary.
    }
    \label{fig:conditional_generation}
\end{figure}

Based on our analysis in Section 4.1, we can identify the contribution of each resolution over time steps. This observation suggests a way to control the influence of conditions on specific resolutions through time-dependent conditioning in conditional generation tasks such as text-to-image. Here, we propose, as an example, a time-dependent prompt composing that enables us to modulate the degree of influence each prompt has on each resolution. 

Given text-to-image diffusion models, we consider a scenario where we aim to generate a city skyline image with the texture of an oil painting. The most naive approach is to generate it with a prompt ``oil painting of a skyline." However, as we can see from the left side of Figure~\ref{fig:conditional_generation}, this method not only creates the texture of an oil painting but also results in the image being overall flat and simplistic. This is because the information from the ``oil painting" contributes not only to texture generation but also affects the coarse features of the image.

Now we consider multiple conditions, $c_i$, as depicted in Equation \eqref{eq:eps_composed}.
Following the proposal by \citet{liu2022compositional}, employing multiple prompts with time-dependent characteristics can control the impact of each prompt on each resolution. This is achived by adjusting the temporal weight $w_i(t)$ associated with each condition.

Returning to the problem of generating a skyline image with the texture of an oil painting, we can solve this by constructing the skyline at a low resolution corresponding to the coarse features, and inserting the oil painting prompt at a high resolution corresponding to the texture information. Here, we chose the simplest function that switches prompts at a specific point in time. Namely,
\begin{equation}
    \begin{aligned}
        w_1(t) &= 1 - H(t- \eta T),  \\
        w_2(t) &= H(t- \eta T),
    \end{aligned}
\end{equation}
where $H(\cdot)$ denotes the Heaviside step function and $\eta$ is a resolution control parameter.

Figure~\ref{fig:conditional_generation} demonstrates the results of generation through time-dependent prompt combining. In this experiment, the prompt starts with ``A HD photo of $\{*\}$" and switches to ``An oil painting of $\{*\}$," with $\eta$ controls the timing of this switch. According to the results, the overall structure of the photo begins to change at $\eta=0.8$, with only minor changes being observed thereafter. The resolution chromatography presented in the top of Figure \ref{fig:conditional_generation} clarifies this phenomenon; at $t=800$, the signals for lower than $8\times8$ resolution are nearly complete, and subsequent changes contribute only minor details, such as the texture of the oil painting, for resolutions higher than $16\times16$. Additional examples can be found in Appendix D. 

\section{Conclusion}

In the generation process of diffusion models, coarse features typically manifest in the early stages, followed by the emergence of detailed features later. Despite this frequent pattern, a complete understanding of this phenomenon remains elusive.
In our study, we identified a scaling relation among samples across various resolutions, each exhibiting distinct signal-to-noise ratios and intensities. Consequently, transforming between these resolutions necessitates time adjustments and intensity rescaling during the generation process.
Then, the scaling relation provides the concept of {\it resolution chromatography}, which represents the relative signal generation rate of each resolution.

Resolution chromatography could contribute to comprehending and implementing diffusion models across various aspects.
First, it offers a mathematical understanding of the coarse-to-fine behavior in the generation process, enabling the quantification of which resolution signals are predominantly generated at any given time.
This helps in deciphering previously explored techniques that involve time-dependent manipulations during the sampling process.
Second, as resolution chromatography is dictated by the noise schedule, reevaluating the role of noise schedules could provide fresh insights. This reexamination could enhance the design of noise schedules, which, thus far, have been predominantly guided by heuristics.
Finally, it suggests an idea for upscaling the resolution of pre-trained models.

Our experiments validated the consistency of resolution chromatography in text-to-image models with our theoretical understanding. Furthermore, when integrating low-resolution models with high-resolution residual models within the cascaded diffusion model, we confirmed the essentiality of employing the scaling relation for time adjustment and intensity rescaling. Additionally, we introduced time-dependent prompt composing as a fundamental example of temporal manipulation for text-to-image diffusion models, demonstrating its efficacy in controlling the resolution of generated images.

However, this study still exhibits certain limitations and ample room for improvement. 
While the theoretical resolution chromatography, derived from the provided noise schedule, generally aligned with the experimental chromatography of CFG, discrepancies in specific details were observed. These differences could stem from the uneven distribution of signal within the image dataset across the frequency domain, as well as the incomplete training of the backward process to accurately replicate the trajectory of the forward process. Essentially, the resolution chromatography of CFG functioned as an indirect validation approach for our theory. However, we still need to develop a direct measurement method for resolution chromatography in image generation.

The potential for applied research leveraging resolution chromatography is vast. In our experiments, we employed the simplest form of the Heaviside step function to compose prompts, but this could be substituted with more sophisticated functions. Going beyond text prompts, a multitude of options exist for temporal manipulation. For example, although stemming from a different context, \citet{kwon2022diffusion} introduced semantic image editing using specialized time-dependent weights, and our theory helps understand how such manipulations contribute to various resolutions. Additionally, a widely used image editing technique known as stochastic differential editing (SDEdit; \cite{meng2021sdedit}), which involves adding noise to an input image and then denoising it, is significantly influenced by the timing of applying forward diffusion. Referencing chromatography enables a clearer understanding of how timing influences the desired level of resolution modification.

Moreover, we expect resolution chromatography becoming a cornerstone for future research endeavors in the design and analysis of noise schedules. Despite numerous studies highlighting the significant influence of noise schedules on generation quality, an effective theoretical framework for their design remains elusive. Accounting for the dataset's power spectrum characteristics alongside chromatography could enhance the design of noise schedules.

\acks{
This work was supported by the Creative-Pioneering Researchers Program through Seoul National University, and the National Research Foundation of Korea (NRF) grant (Grant No. 2022R1A2C1006871).
}

\vskip 0.2in
\bibliography{sample}

\newpage
\appendix
\section{Theoretical Derivation of Resolution Chromatography}
\subsection{Proof of Theorem \ref{thm:equiv_of_chrom}}
Let us define a remapped time $t'(t)$ that ensures $\alpha_t=\alpha'_{t'}$ for two monotonically decreasing noise schedules $\alpha_t$ and $\alpha'_t$ concerning time $t$. Consequently, the adjusted times for $t$ and $t'$ in the $m$-th downsampling are $\tau_m$ and $\tau'_m$, respectively, leading to the corollary $\alpha_{\tau_m}=\alpha'_{\tau'_m}$.

Then, given the definition in Equation \eqref{eq:def_chromatography}, the resolution chromatography of $\alpha'_t$ measured on the original time $t$ becomes
\begin{equation}
    r'_m(t) = \left( \sum_n \frac{d\alpha'_{\tau_n}}{dt} \right) ^{-1} \frac{d\alpha'_{\tau_m}}{dt}.
\end{equation}
To derive the relation $r'_m(t')=r_m(t)$ in the theorem, we explicitly state the remapped chromatography to be
\begin{align}
    r'_m(t') &= r'_m(t=t') \nonumber \\
    &= \left( \sum_n \frac{d\alpha'_{\tau_n}}{dt} \bigg|_{t=t'} \right) ^{-1} \frac{d\alpha'_{\tau_m}}{dt} \bigg|_{t=t'} \nonumber \\
    &= \left( \sum_n \frac{d\alpha'_{\tau'_n}}{dt'}  \right) ^{-1} \frac{d\alpha'_{\tau'_m}}{dt'} \nonumber \\
    &= \left( \sum_n \frac{d\alpha'_{\tau'_n}}{dt} \frac{dt}{dt'}  \right) ^{-1} \frac{d\alpha'_{\tau'_m}}{dt} \frac{dt}{dt'}\nonumber \\
    &= \left( \sum_n \frac{d\alpha_{\tau_n}}{dt} \frac{dt}{dt'}  \right) ^{-1} \frac{d\alpha_{\tau_m}}{dt} \frac{dt}{dt'}\nonumber \\
    &= \left( \sum_n \frac{d\alpha_{\tau_n}}{dt}  \right) ^{-1} \frac{d\alpha_{\tau_m}}{dt} = r_m(t).
\end{align}
This formulation offers a convenient method for computing resolution chromatography. In various studies, the variable $t$ is assigned different ranges, such as $[0, T]$, $[0, 1]$, and $[0, \infty)$, creating challenges in comparing ${d\alpha_{\tau_m}}/{dt}$ due to the distinct time scales in each scenario. However, the normalized chromatography $r_m(t)$ remains invariant to the time scale, enabling comparisons across these cases.

\subsection{Natural Noise Schedule}
Here, we propose the \textit{natural noise schedule} as a reference. We consider the Ornstein-Uhlenbeck process as the natural noise schedule that is defined by the following stochastic differential equation:
\begin{equation}
    dx_t = -\theta x_t + \sigma dW_t,
\end{equation}
where $\theta$ and $\sigma$ are drift and diffusion parameters,
and $W_t$ denotes the Wiener process ~\citep{jacobs2010stochastic}. Although it takes infinite time to converge to a Gaussian distribution, this equation represents physical diffusion (Brownian motion) in a quadratic potential, making it \textit{natural}. The analytic solution of this equation is well known and the mean and variance of a sample $x_0$ over time are
\begin{equation}
    \mathbb{E}[x_t] = x_0 e^{-\theta t}, \quad \mathrm{Var} [x_t] = \frac{\sigma^2}{2\theta} \left( 1 - e^{-2\theta t}\right).
\end{equation}
To make it consistent with Equation \eqref{eq_diffusion}, we set the parameters as $\sigma=1$, $\theta=0.5$ and $t\in [0, \infty)$. 
Then the natural noise schedule becomes
\begin{equation}
    \alpha_t = e^{-t}.    
\end{equation}
This setting satisfies
\begin{align}
    x_t &= \sqrt{\alpha_t}x_0 + \sqrt{1-\alpha_t}\epsilon_t \nonumber \\
    &= x_0e^{-t/2}+\sqrt{1-e^{-t}} \epsilon_t.
\end{align}
Then, SNR $=1/{(e^t -1)}$, and its inverse function is $t(\textrm{SNR})=\ln (1+ \textrm{SNR}^{-1})$. 
Using Equation \eqref{eq:generalized_relation}, the corresponding time adjustment can be derived as follows:
\begin{align}
\label{eq:tau_m}
\tau_{m} &=\ln \left(1+\frac{1}{2^{2m}\textrm{SNR}(t)}\right) \nonumber \\
&= \ln(e^t+4^m-1) - m\ln 4.
\end{align}
When $t$ is sufficiently large, $\tau_{m+1} \approx \tau_m-\ln4$.
This result represents that the time adjustment under downsampling can be approximated as a constant time shift in the natural noise schedule. 
Furthermore, using Equation \eqref{eq:tau_m}, we can obtain the resolution chromatography of the natural noise schedule,
\begin{equation}
\label{eq_natural_chromatography}
    \begin{aligned}
        r_m(t) = \frac 1 Z \frac{d\alpha_{\tau_m}}{d{\tau_m}}\frac{d\tau_m}{dt} = -\frac 1 Z 
        \frac{e^{-\tau_m}}{(4^m-1)e^{-t}+1}.
    \end{aligned}
\end{equation}
Note that $d\alpha_t/dt < 0$ because SNR decreases in forward diffusion, but $r_m(t)$ becomes positive due to the negative normalization constant $Z =\sum_m d \alpha_{\tau_m}/dt < 0$.

\subsection{Resolution Chromatography for Arbitrary Noise Schedules}
Let $\alpha_t^*$ denote the natural noise schedule and $\alpha_t$ denote an arbitrary noise schedule. According to Theorem~\ref{thm:equiv_of_chrom}, the remapped time $t^*(t)$, that satisfies $\alpha_{t^*}^*=\alpha_t$, is all we need to get the chromatography. By using the definition of the natural schedule, $e^{-t^*}=\alpha_t$, or $t^*=-\ln \alpha_t$, we can obtain the chromatography in an analytical way.

Rewriting the natural chromatography in Equation \eqref{eq_natural_chromatography}, we can directly calculate $r_m(t)$ of any noise schedule using Theorem \ref{thm:equiv_of_chrom}:
\begin{align}
    r_m(t) = r^*_m(t^*) &= \left(\sum_{n} {\frac{4^{n} e^{t^*}}{(e^{t^*}+4^{n}-1)^2}}\right)^{-1} \frac{4^m e^{t^*}}{(e^{t^*}+4^m-1)^2} \\
    &= r^*_m(-\ln \alpha_t).
\end{align}
It is interesting that the form of $r_m^*({{t^*}})$, leaving aside its scale, is similar to the derivative of sigmoid function, $d\sigma(x)/dx = e^x/(e^x+1)^2$. This explains why $r_m(t)$ is bell-shaped.

To obtain the chromatography for a specific noise schedule, we simply calculate the natural chromatography $r^*_m$ and just read its value at ${t^*}=-\ln \alpha_t$. For example, if we use the cosine noise schedule, $t^*$ would be
\begin{equation}
    t^*=-\ln \cos \left( \frac{t/T+s}{1+s} \cdot \frac \pi 2 \right)^2 + \ln \cos \left( \frac{s}{1+s} \cdot \frac \pi 2 \right)^2.
\end{equation}

\section{Static Threshold for Multiple Resolutions}
In our experiment, to improve image quality, we applied a static threshold \citep{saharia2022photorealistic} due to the pixel range limitation in $x_0$ being constrained between $[-1, 1]$. However, our method differs from conventional approaches that utilize a single noise predictor; instead, we use the sum of noise predictors from multiple resolutions. Consequently, certain adjustments are required to apply the static threshold to each resolution individually.

By appropriately adjusting for down-sampling, the following is derived from Equation \eqref{eq:x0_expectation}:
\begin{equation}
    \mathbb{E}\left[x_0^{(m)}|x_{\tau_m}^{(m)}\right]=\frac{1}{\sqrt{\alpha_{\tau_m}}} \left[ \lambda_t^{(m)}\D^m[x_t] - 2^m \sqrt{1-\alpha_{\tau_m}} \D^m[\epsilon_t] \right].
\end{equation}
The procedure involves applying a static threshold to $\mathbb{E}\left[x_0^{(m)}|x_{\tau_m}^{(m)}\right]$, derived from each resolution model, and subsequently summing them as outlined in Algorithm 1.

\begin{algorithm}
\caption{Static Threshold for Multiple Resolutions} \label{alg:casdaded}
\begin{algorithmic}
\Require Sample $x_t$, predicted noise $\epsilon_t$, threshold value $q$ \\
\Function{\texttt{getResidual}}{$x$}
    \State \Return $x - \U[\D[x]]$
\EndFunction \\
\Function{\texttt{MultiresolutionThresholding}}{$x, q$}
    \State $M \gets \lfloor \log_2(\text{size of } x_0) \rfloor$ \Comment{Max cascading number}
    \State $x_0 \gets \mathbf{0}$
    \For{$m = M$ \textbf{to} $0$}
        \State $\alpha_{\tau_m} \gets \frac{2^{2m} \alpha_t} {(2^{2m}-1)\alpha_t + 1}$
        \State $\lambda_t^{(m)}\gets \frac{2^m}{\sqrt{1 + (2^{2m}-1) \alpha_t}}, \phantom{M}$
        \State $x_0^{(m)} \gets \frac{1}{\sqrt{\alpha_{\tau_m}}} \left( \lambda_t^{(m)} \D^m[x_t] - 2^m \sqrt{1-\alpha_{\tau_m}} \D^m[\epsilon_t] \right)$
        \If{$m>0$}
            \State $x_0 \gets x_0 + \texttt{resize} \left( \texttt{getResidual}(x_0^{(m)}), \text{size} = x_0.\text{size} \right)$
        \Else
            \State $x_0 \gets \texttt{resize} \left( x_0^{(m)}, \text{size} = x_0.\text{size} \right)$
        \EndIf
        \State $x_0 \gets x_0.\texttt{clamp}(\text{min}=-1, \text{max}=1)$ \Comment {Static threshold}
    \EndFor
    \State \Return $x_0$
\EndFunction
\end{algorithmic}
\end{algorithm}

\newpage
\section{More Examples of Measured Resolution Chromatography}
\begin{figure}[h]
    \centering
    \includegraphics[width=0.8\linewidth]{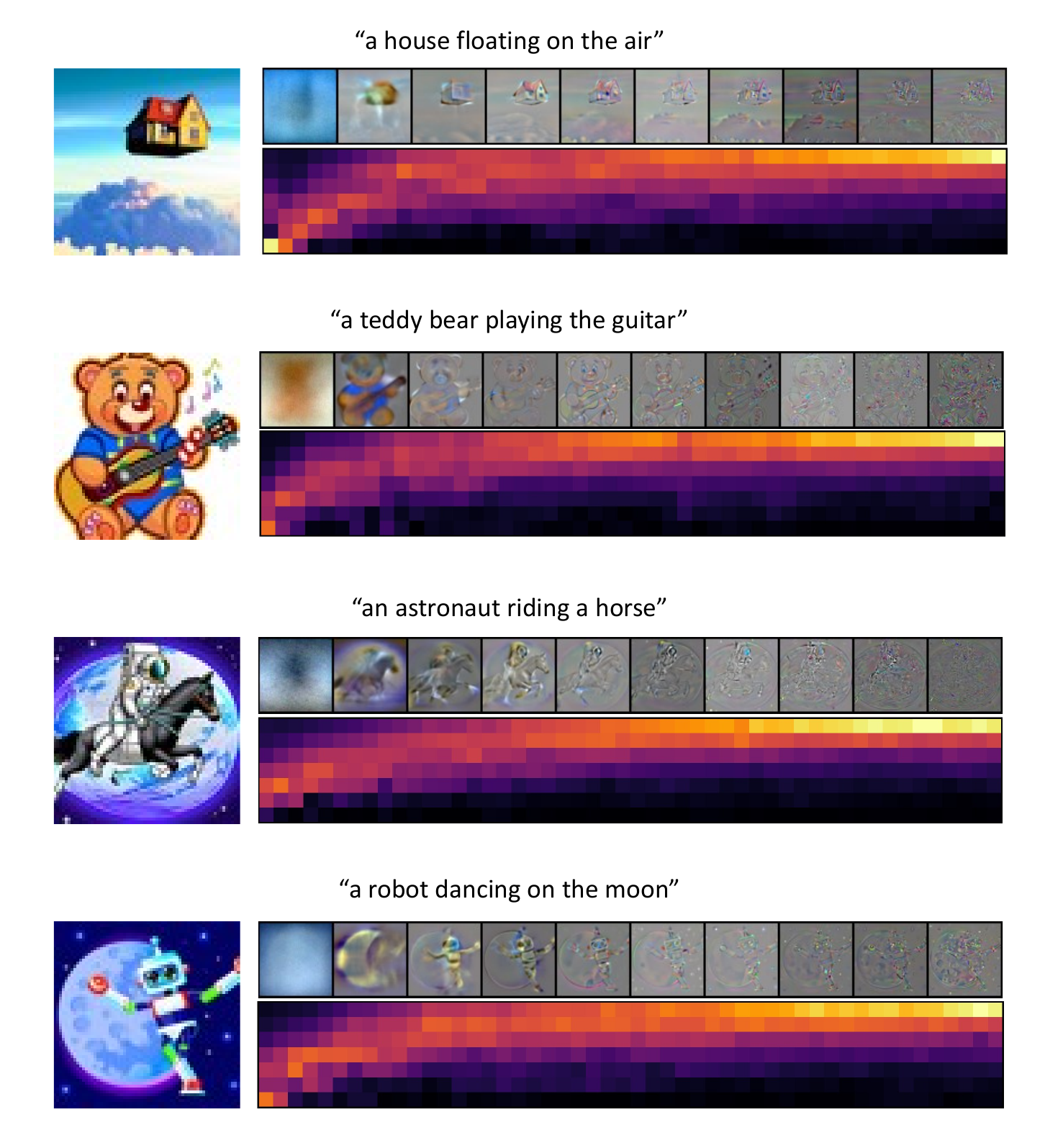}
    \caption{Resolution chromatography of CFG diffusion models. Samples are generated from various text prompts and their classifier-free guidance terms. The heat map represents the resolution chromatography measured by the method introduced in Section 4.1.}
    \label{fig:cfg_examples}
\end{figure}

\newpage
\section{More Examples of Time-Dependent Prompts Combining}

\begin{figure}[H]
    \centering
    \includegraphics[width=0.8\linewidth]{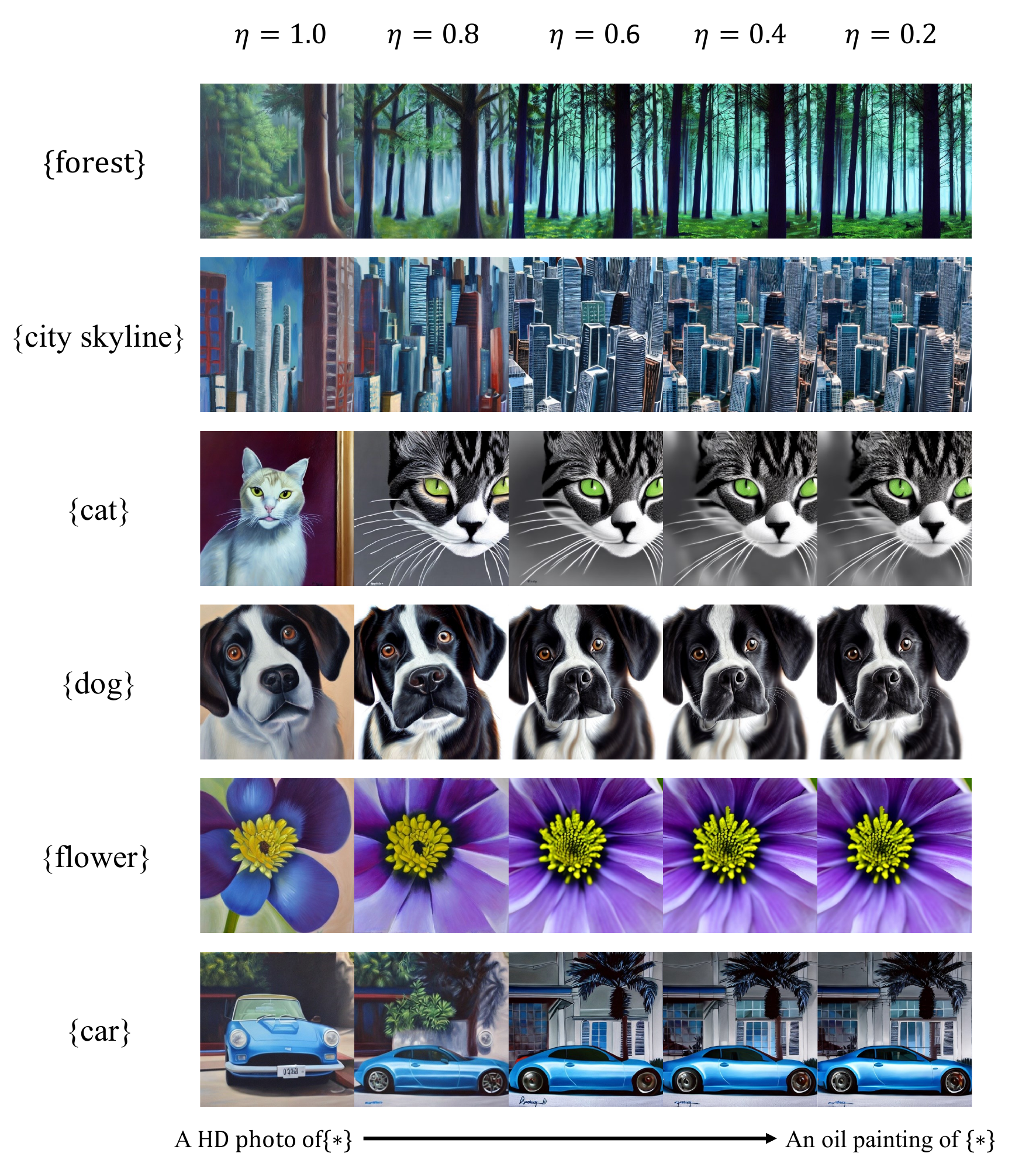}
    \caption{Time-dependent conditional image generation. 
    In all examples, a rapid change in the coarse feature occurs at $\eta=0.8$, which is an anticipated outcome derived from the theoretical resolution chromatography in Figure \ref{fig:cfg}. } 
    \label{fig:conditional_generation_more}
\end{figure}

\end{document}